\documentclass[10pt,twocolumn,letterpaper]{article}

\usepackage{cvpr}
\usepackage{times}
\usepackage{epsfig}
\usepackage{graphicx}
\usepackage{amsmath}
\usepackage{amssymb}
\usepackage{multirow}
\usepackage[ruled,vlined]{algorithm2e}
\DeclareMathOperator*{\argmin}{argmin}

\usepackage[pagebackref=true,breaklinks=true,letterpaper=true,colorlinks,bookmarks=false]{hyperref}

\cvprfinalcopy 

\def\cvprPaperID{****} 

\ifcvprfinal\fi
\begin{document}
	
	\title{Rotation Averaging with Attention Graph Neural Networks}
	
	\author{Joshua Thorpe\\
		School of Engineering\\
		RMIT University Australia\\
		{\tt\small josh.thorpe@rmit.edu.au}
		\and
		Ruwan Tennakoon\\
		School of Science\\
		RMIT University Australia\\
		{\tt\small ruwan.tennakoon@rmit.edu.au}
		\and
		Alireza Bab-Hadiashar\\
		School of Engineering\\
		RMIT University Australia\\
		{\tt\small abh@rmit.edu.au}
	}
	
	\maketitle
	
	\begin{abstract}
		In this paper we propose a real-time and robust solution to large-scale multiple rotation averaging.
        Until recently, Multiple rotation averaging problem had been solved using conventional iterative optimization algorithms. Such methods employed robust cost functions that where chosen based on assumptions made about the sensor noise and outlier distribution.
        In practice, these assumptions do not always fit real datasets very well.
        A recent work showed that the noise distribution could be learnt using a graph neural network. This solution required a second network for outlier detection and removal as the averaging network was sensitive to a poor initialization.
        In this paper we propose a single-stage graph neural network that can robustly perform rotation averaging in the presence of noise and outliers. Our method uses all observations, suppressing outliers effects through the use of weighted averaging and an attention mechanism within the network design.
        The result is a network that is faster, more robust and can be trained with less samples than the previous neural approach, ultimately outperforming conventional iterative algorithms in accuracy and in inference times. 
	\end{abstract}
	
	\section{Introduction}
	Rotation averaging is a computer vision problem that has gained significant attention in recent years. It is a common geometric problem that presents itself in structure from motion (SFM), bundle adjustment, multiple camera calibration and sensor network calibration problems. 
	Rotation averaging aims to estimate a set of absolute camera orientations $\{R^*_i\}$ given a set of noisy relative rotations between cameras $\{R_{ij}\}$. This is a high dimensional non-linear optimization problem that has been optimally solved using a Bundle adjustment method \cite{Hartley2013}; however, the overwhelming computational cost of this process has motivated faster algorithms to be developed. These faster optimization algorithms use efficient optimizers that have allowed rotation averaging to be applied to increasingly larger datasets. 
	
	Deep learning has been driving state-of-the-art solutions in many computer vision problems in recent times. Pulak et. al. \cite{Purkait2019} identified that the noise distribution in the relative rotations between cameras was inconsistent between datasets and often not well captured by the cost functions used with these efficient optimization algorithms, and so created the first neural based approach to rotation averaging that generally outperformed all previous optimization methods. The benefits of using a deep learning approach is that the noise distribution is learnt rather than tuned for, providing more accurate and robust predictions of absolute camera orientations. Modern deep learning frameworks allow for inference times that are many orders of magnitude faster than the traditional iterative optimization algorithms, opening up the possibility of real-time rotation averaging, useful for calibrating rigs with hundreds and potentially thousands of cameras. 
	This neural approach used two sequential networks. The first network cleans the set of relative rotations and removes observations from the network it believes to be outliers. An initial guess of each cameras orientation is achieved using a spanning tree algorithm before the second network performs rotation averaging to produce refined predictions for each cameras absolute orientation. The multistage sequential operation is cumbersome and slow to train with little insight to the performance of each network individually. The complexity of the approach affects the practicality of implementing the solution and slows inference times significantly when you account for network initialization times and the spanning tree operation required between each network inference.
	
	In this paper we address these problems, outlining a single neural network approach to rotation averaging that is less convoluted, gives faster inference and training times, and outperforms the previous methods. We describe a graph neural network based on the message passing framework that utilizes edge attention mechanisms to robustly overcome outliers in the input set of relative rotations without the need for a preliminary cleaning network.
	
	
	\section{Multiple Rotation averaging}
	The multiple rotation averaging (MRA) problem can be described as computing the absolute orientations of several cameras in a common global frame, given a set of noisy relative orientations between the cameras.\\
	The relationship between cameras can be represented by a viewgraph $G=\{V,\varepsilon\}$, where the vertices $V$ represents each camera and the edges $\varepsilon$ corresponding to the relative relationship between cameras. $V_v \in V$ is represented by the absolute rotation $R_v$ and $\varepsilon_{uv} \in \varepsilon$ is represented by the relative rotation $R_{uv}$.
	All rotations belong to the 3D rotation group $R \in SO(3)$ and satisfy the compatibility constraint $R_{uv}=R_vR_u^{-1}$
	
	Given a set of noisy relative camera orientation observations $\{\tilde{R}_{uv}\}$, the conventional optimization problem to solve for the set of predicted absolute camera orientations $\{R^*_v\}$can be described as 
	
	\begin{equation}\label{eq:optimization_MRA}
		\argmin_{\substack{\{R^*_v\}}}\sum_{\varepsilon_{uv}\in\varepsilon}\rho(d(\tilde{R}_{uv},R^*_v{R^*_u}^{-1}))
	\end{equation}

	given $\rho(.)$ is a robust cost function (discussed in \ref{sec:robust_cost}) and $d(.)$ is a distance measure between two rotations (described in \ref{sec:distance_measure}). See table \ref{tab:nomenclature} for nomenclature used throughout this manuscript.

	\subsection{Distance Measures}\label{sec:distance_measure}
	The three commonly used distance measures are the geodesic or angular distance $d_\theta=\angle(\tilde{R},\hat{R})$, the chordal distance $d_C=||\tilde{R}-\hat{R}||_F$, and the quarternion distance $d_Q=\min\{||q_{\tilde{R}}-q_{\hat{R}}||,||q_{\tilde{R}}+q_{\hat{R}}||\}$, where $||.||_F$ is the Frobonius norm, while $q_{\tilde{R}}$ and $q_{\hat{R}}$ are the quarternion representations of rotations $\tilde{R}$ and $\hat{R}$ respectively. It was proven in \cite{Hartley2013} that the chordal and quarternion are related to the geodesic distance by $d_C=2\sqrt{2}\sin(d_\theta/2)$ and $d_Q=2\sin(d_\theta/4)$ respectively. The quarternion representation is used to represent all rotations in the proposed network due to its computational efficiency as a 1D vector and so the quarternion distance measure is chosen when calculating error in the loss function. The final graph alignment and performance metrics utilize the geodesic measure, as angular error is a scalar that is more intuitive and relatable.
	
	\subsection{Robust cost functions}\label{sec:robust_cost}
	Various robust cost functions have been used to overcome sensor and environmental noise that appears in various rotation averaging applications.
	Outliers are usually created by environmental factors such as repeated identical features of man-structures that result in corrupted relative rotations being generated.
	To overcome noise and outliers, conventional methods optimize \ref{eq:optimization_MRA} with a robust cost function $\rho(.)$ \\
	\cite{Chatterjee2017} provides an extensive comparison of all the robust cost functions used in conventional optimization methods.	
	These robust cost functions are designed around assumptions made about sensor noise and outlier distributions. \cite{Purkait2019} observed that real datasets do not have noise and outlier distributions that match the assumptions made by previous works when choosing their respective robust cost functions. This is the motivation behind using deep learning methods, to learn the noise and outlier distribution, outperforming any conventional robust optimization algorithm.
	
	\begin{table}[t]
		\small
		\begin{center}
			\begin{tabular}{|ll|ll|}
				\hline
				\multicolumn{4}{|c|}{Rotation parameters for the cameras in the viewgraph} \\
				\hline
				$\Tilde{R}_{uv}$&   :Input relative&        $\Tilde{R}_v$&  :SPT Initialized absolute\\
				$\hat{R}_{uv}$&     :Ground truth relative& $\hat{R}_v$&    :Ground truth absolute\\
				&                   &                       $R^*_v$&        :Predicted absolute\\
				\hline
				\multicolumn{4}{|c|}{Network functions and parameters} \\
				\hline
				\multicolumn{2}{|l|}{$f_m, f_u, f_a, f_{ro}$}& \multicolumn{2}{|l|}{: Multi-layer perceptron networks} \\
				\hline
				
			\end{tabular}
		\end{center}
		\caption{Nomenclature used in this paper}
		\label{tab:nomenclature}
	\end{table}
	
	
	\section{Existing Methods}
	Two unique approaches have been explored, conventional optimization for MRA and deep learning based methods.
	
	Govindu was the pioneer of conventional optimization for MRA with his linear motion model method \cite{Govindu2001} that exploited the relation between cameras relative algebraic constraints and a global motion model.
	Govindu later improved his method with a faster proposal \cite{Govindu2004} that utilised Lie-group based averging.
	Both solutions were susceptible to outliers, and so outlier detection methods were later used by \cite{Govindu2006} and \cite{Zach2010} to remove outliers prior to averaging. \cite{Govindu2006} used a RANSAC method to detect and remove outliers whilst \cite{Zach2010} used the fact that a loop of rotations should result in the identity rotation if no noise or outliers exist in the loop, to identify and remove outliers. Both these methods to identify and eliminate outliers are very computationally expensive, especially as the number of cameras gets large.
	
	Robust optimization algorithms were later used to suppress the influence of outliers, allowing good results whilst including all observations. This removed the computationally costly step of removing outliers.
	
	\cite{Crandall2011} proposed a method they called DISCO. A two stage process that performs a rough initialization followed by a refinement. The first stage of DISCO reduces the rotation space from SO(3), by ignoring the twist component of rotations and then employs a loopy belief propagation to solve the discrete optimization. The second stage applies continuous Levenberg-Marquardt refinement to fine-tune the predicted orientations.
	
	\cite{Hartley2011} introduced a single stage averaging method by solving the minimization optimization problem using the Weiszfeld algorithm directly on the SO(3) group. The $\ell_{1}$ mean employed in the algorithm is more robust than previously used $\ell_{2}$ norm, resulting in each cameras absolute orientation being updated with the median of the neighbors predictions.
	
	\cite{Chatterjee2017} also employed a two-stage process of performing a rough averaging to improve the initialization before running a refining stage to fine-tune the predicted orientations. They employ an iterative $\ell_{1}$ loss minimization for the roughing stage, followed by the more robust $\ell_{\frac{1}{2}}$ loss minimization for fine-tuning.\\
	
	When it comes to learning based methods, \cite{Purkait2019} is the only deep learning work proposing a solution to the problem of MRA with their model they call NeuRoRA. NeuRoRA is a two stage process, each including its own graph neural network. The first stage has a neural network that refines the observed relative rotations and removes outliers, with the aim to improve the spanning tree initialization that gives each camera an initial absolute orientation. The second network then performs the averaging, refining the absolute orientation predictions for each camera. The first network learns the noise and outlier distribution of the relative rotations and provides an edge level output with refined relative rotations and an outlier score for each edge. The second network learns the noise distribution of the relative rotations and provides a node level output which is in the form of a rotation correction factor that is multiplied to each node's initialization orientation in order to refine it.\\
	
	Each method has a major limitation that is still driving research in MRA. The early conventional methods \cite{Govindu2001}, \cite{Govindu2004}, \cite{Govindu2006} and \cite{Zach2010} as well as the neural method proposed by \cite{Purkait2019} are all performing non-robust averaging that is susceptible to outliers. \cite{Govindu2006}, \cite{Zach2010} and \cite{Purkait2019} had to employ a pre-cleaning stage to clean the data from any outliers to overcome this issue. This can be extremely computationally costly in the cases of \cite{Govindu2006} and \cite{Zach2010} but as discussed by \cite{Chatterjee2017}, any approach that identifies and removes outliers, requires a specific threshold to be set. A simple threshold does not correctly classify outliers in favorable conditions let alone in real datasets, and so making the assumption about the outlier distribution and attempting to remove outliers is often computationally expensive and impractical.
	
	The robust optimization methods such as \cite{Hartley2011} and  \cite{Chatterjee2017}, although faster and more robust than previous methods whilst utilizing all the observations still make assumptions about the noise and outlier distribution based on the choice of robust cost function. 
	Lastly, methods such as  \cite{Chatterjee2017} praise themselves on being significantly computationally more efficient than previous conventional optimization problems but are still slow on large datasets as highlighted by the neural learning based method \cite{Purkait2019} which is many orders of magnitude faster again. NeuRoRA is the only method that is approaching inference times capable of real-time MRA thanks to the use of highly parallelized deep learning frameworks with GPU acceleration.
	
	NeuRoRA's neural approach is potentially revolutionary for MRA with its claimed ability to learn the noise distribution of data without assumptions being made. Its biggest limitation is that its averaging network is not robust to outliers, requiring a complex setup of two sequential networks, the first of which classifies and removes outliers based on pre-defined assumptions.

	\section{Proposed Method}
	\subsection{Network Framework}
	The proposed network is based on a message-passing neural network (MPNN) framework introduced by \cite{Gilmer2017}. Given a view-graph $G$ made up of nodes $V$ and edges $\varepsilon$, the MPNN can generate any combination of node, edge or graph level predictions through the use of a message passing phase and a readout phase. The message passing phase is defined by a message function $f_m$ and an update function $f_u$. During this phase the hidden representation of a node $h_v^{\left \langle t \right \rangle}$ is updated based on the messages $M_v^{\left \langle t \right \rangle}$ propagating from neighboring nodes over $T$ timesteps according to
	
	\begin{equation}
		M_v^{\left \langle t \right \rangle}= \square_{u\in \mathcal{N}_v} f_m(h_v^{\left \langle t-1 \right \rangle},h_u^{\left \langle t-1 \right \rangle},\epsilon_{uv}^{\left \langle t-1 \right \rangle})
	\end{equation}

	\begin{equation}
		h_v^{\left \langle t \right \rangle}=f_u(h_v^{\left \langle t-1 \right \rangle}, M_v^{\left \langle t \right \rangle})
	\end{equation}
	
	where $\mathcal{N}_v$ is the set of all neighbors to $V_v$ in graph $G$ and $\epsilon_{uv}$ is the edge value between nodes $u$ and $v$. The readout phase involves a readout function $f_{ro}$, which for a node level output would be
	\begin{equation}
		\hat{y}_v = f_{ro} (h_v^{\left \langle T \right \rangle})
	\end{equation}
	
	$f_m$ and $f_u$ and $f_{ro}$ are all learnable differentiable functions.
	
	\subsection{Network Design for MRA}
	The MPNN framework was intended to have the hidden values $h_v^{\left \langle t \right \rangle}$ updated several times over $T$ timesteps before a node, edge or graph level output be generated using the readout function. For MRA applications, completing multiple iterations of message passing operations using abstract high-dimensional feature vectors as the hidden representation of the nodes is not beneficial.\\
	The nodes need to represent an SO(3) rotation at all times which can be done with a 4 element quaternion so we limit our network to a single message passing operation before a readout. The second alteration is that when a nodes value is being updated, in MRA there is no need to consider what the nodes value on the previous iteration was, and so the update equation is not used either. 
	
	Given a view-graph $G$ made up of nodes $V$ and edges $\varepsilon$, where the nodes represent a set of absolute rotations $\{R\}$, the network design choice used to predict a refined absolute rotation for node $R_v$ can described as  
	
	\begin{equation}
		R_v^{\left \langle t \right \rangle}=R_v^{\left \langle t-1 \right \rangle}*f_{ro}(M_v^{\left \langle t \right \rangle})
	\end{equation}

	\begin{equation}\label{eq:hidden_update}
		M_{v}^{\left \langle t \right \rangle} =  \sum_{u \in \mathcal{N}_v } {w}_{uv}^{\left \langle t \right \rangle} f_m\left( R_v^{\left \langle t-1 \right \rangle}, R_u^{\left \langle t-1 \right \rangle}, e_{uv}^{\left \langle t-1 \right \rangle} \right)
	\end{equation}

	where $w_{uv}$ is an edge weighting described in equation \ref{eq:w_uv}, $f_{ro}$ and $f_m$ are learnable functions made up of layers of linear perceptrons. $f_{ro}$ is the readout function that converts the hidden representation into a predicted rotation correction that refines the nodes original absolute orientation.  $f_m$ is the message function used in the MPNN framework and produces a prediction for the new $h_v^{\left \langle t \right \rangle}$ for each edge connected to $V_v$.\\
	
	The loss for this network is the combination of two loss functions. The first takes into account the error in the relative rotations between the predicted nodes and the second calculates the error in predicted nodes absolute positions.
	\begin{equation}
	\mathcal{L} = \frac{1}{\left | V \right |} \sum_{v \in V} \left (\frac{1}{\left |\mathcal{N}_v  \right |}\sum_{u \in \mathcal{N}_v} \mathcal{L}_{r}\left( v, u\right) + {\beta} \mathcal{L}_{a}\left( v\right)  \right ) 
	\end{equation}
	where $\mathcal{L}_r$ is the loss for the relative consistency between nodes in the prediction as described in \ref{eq:lr}. $\mathcal{L}_a$ is the loss for the predicted absolute rotations and can be described by \ref{eq:la}. $\beta$ is chosen as 0.25 to reduce the influence of the predicted absolute orientations as we value the consistency of nodes relative to one-another more, since the graph alignment will remove any offset in global frame alignment between the prediction and ground-truth.
	\begin{equation}\label{eq:lr}
		\mathcal{L}_r\left( v, u\right)=d_Q \left ( \hat{R}_v^{-1}\hat{R}_u, {R^*_v}^{-1} R^*_u \right)  \forall (u,v) \in \epsilon
	\end{equation}
	\begin{equation}\label{eq:la}
		\mathcal{L}_{a}\left( v\right) = d_Q \left ( \hat{R}_v, R^*_v\right) \quad \forall v \in V
	\end{equation}
	
	\begin{figure}[t]
		\includegraphics[width=.5\textwidth]{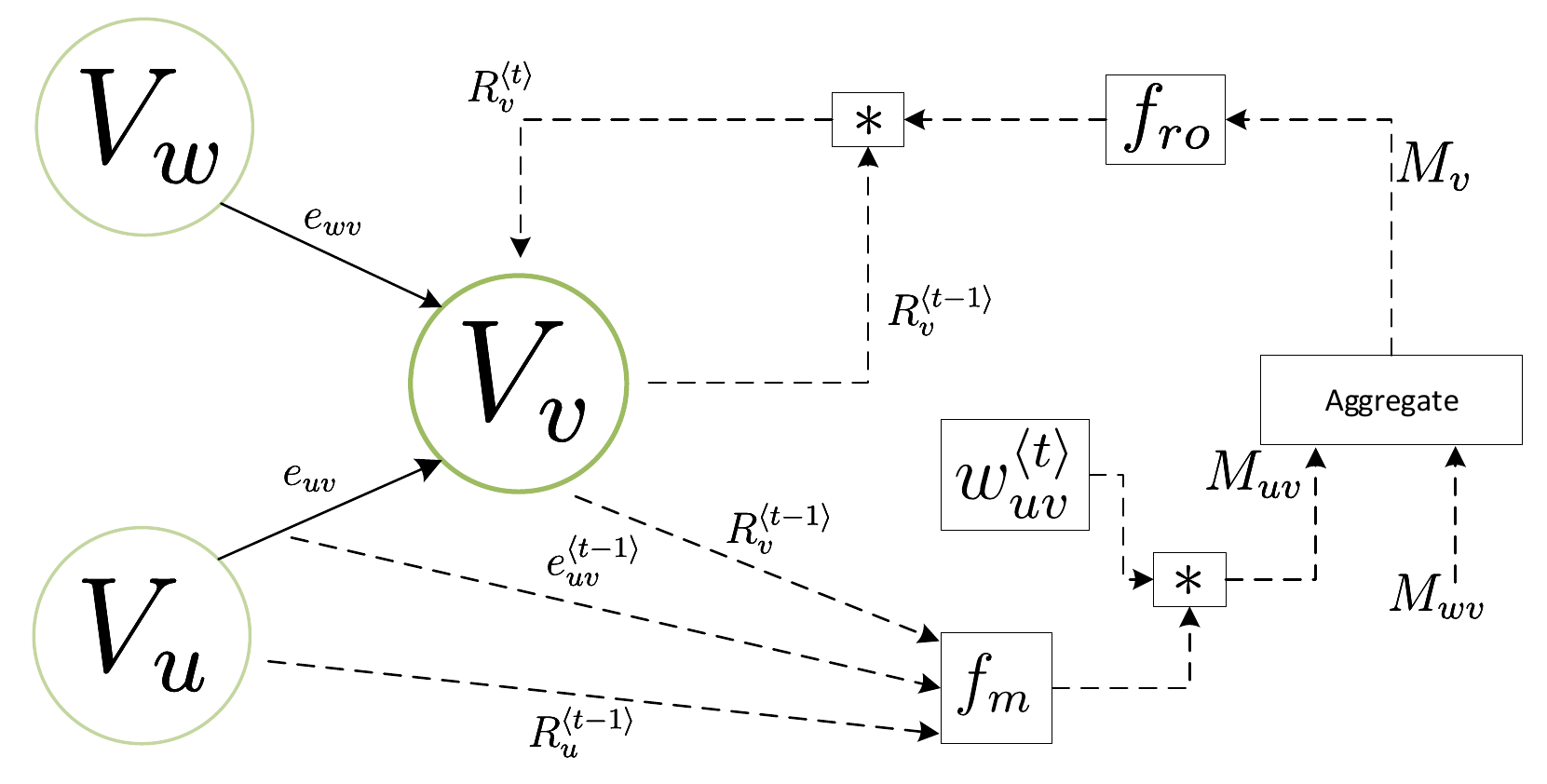}
		\caption{Visual illistration of computing the next timestep node value $R_v$ in the proposed MPNN}
		\label{fig:diagram}
	\end{figure}
		
	\subsection{Network Initialization for Learning Noise}
	In SFM applications of MRA, the input is a set of relative rotations $\{\Tilde{R}_{uv}\}$. The nodes of the graph need to be initialized with a starting value before any MRA operations can be performed. A simple solution proposed by \cite{Hartley2011} is to generate a spanning tree by setting the node (camera) with the maximum number of neighbors as the root node and spanning out, initializing all nodes in the graph by forming a path from the root node without forming a circle. The global reference frame is set at the root node and therefore all other nodes will be initialized with an absolute rotation $\{\Tilde{R}_v\}$ by propergating away from the root using the relative rotations to calculate an absolute rotation for each node, $\Tilde{R}_v=\Tilde{R}_{uv}\Tilde{R}_u$ \\
	The error in the initialized nodes $\{\Tilde{R}_v\}$ is proportional to the noise in the relative rotations $\Tilde{R}_{uv}\hat{R}_{uv}^{-1}$ used in the tree path to each node. Therefore the shortest path tree \cite{Tarjan1982} method that minimizes the sum of the distances of all the initialization paths will minimize the amount of noise and error in the node initializations. The shortest path can be found by running a breadth-first search from the root node. The optimum spanning tree that minimizes the sum of distances of the initialization paths, can be found by applying the breadth-first search with each node as the root node \cite{Hassin1995}. The time complexity of running the spanning tree for each node as the root is $\mathcal{O}(n^2)$. Similar to \cite{Hartley2011} and \cite{Purkait2019} we set the root node to the node with the most neighbors, as it has been shown to produce results comparable to the optimum spanning tree method due to this node often being located at the center of the graph in SFM applications. The time complexity is then reduced to $\mathcal{O}(n)$ and the node initialization can be achieved in comparable times to the rotation averaging inference times. For reference, this method will be referred to as shortest path tree (SPT).\\
	
	The network cannot learn to predict absolute rotations for nodes in a viewgraph. To perform MRA it must learn the noise distribution of the graph as discovered by \cite{Purkait2019}. This is done by calculating the discrepancy between the observed input set of relative rotations $R_{uv} \in \varepsilon$ and the relative rotation between the initial node values $R_v^{-1}R_u$. The set of edge attributes used as input to the network at each timestep can be can therefore be initialized according to 
	
	\begin{equation}
		e_{uv}^{\left \langle 0 \right \rangle} = \Tilde{R}_v^{-1} \Tilde{R}_{uv}  \Tilde{R}_u
	\end{equation}
	
	and the set of nodes attributes in the input graph are set to the node values from the SPT initialization 
	\begin{equation}
		R_v^{\left \langle 0 \right \rangle}=\Tilde{R_v}
	\end{equation}
	
	\begin{algorithm*}[t]
		\SetAlgoLined
		\textbf{Input:} $R_{v}$, $\Tilde{R}_{uv}$, $t_{max}$, $\beta$  \\
		Compute $e_{uv}^{\left \langle 0 \right \rangle} = R_v^{-1} \Tilde{R}_{uv} R_u$ for all edges\;
		$t \gets 1$ \;
		\While{$t \leq t_{max}$}{
			\For{$v \in V$}{
				For all $u \in \mathcal{N}_v$ compute: $m_{uv}^{\left \langle t \right \rangle} = f_m\left( R_v^{\left \langle t-1 \right \rangle}, R_u^{\left \langle t-1 \right \rangle}, e_{uv}^{\left \langle t-1 \right \rangle} \right) \quad \triangleright \textrm{An~MLP}$ \;
				Compute relative neighbourhood size of all $u \in \mathcal{N}_v$: $n_{uv} = \frac{\left |\mathcal{N}_u  \right |}{\max_{\{j \in \mathcal{N}_v\}} |\mathcal{N}_j |}$ \;
				Compute attention weights: $\hat{w}_{uv} = f_a\left( n_{uv}, e_{uv}^{\left \langle t-1 \right \rangle} \right) \quad \triangleright \textrm{An~MLP}$ \;			
				Normalize attention weights: ${w}_{uv}^{\left \langle t \right \rangle} = \frac{\exp \left (\hat{w}_{uv}  \right )}{\sum_{j\in \mathcal{N}_v} \exp \left (\hat{w}_{jv}  \right )}$ \;		
				Compute the aggregated massage: $M_{v}^{\left \langle t \right \rangle} =  \sum_{u \in \mathcal{N}_v } {w}_{uv}^{\left \langle t \right \rangle} m_{uv}^{\left \langle t \right \rangle} $ \;
			}
			For all nodes $v \in V$ predict the new rotations: $R_v^{\left \langle t \right \rangle} =  R_v^{\left \langle t-1 \right \rangle}*f_{ro}\left( M_{v}^{\left \langle t \right \rangle} \right)  \quad \triangleright \textrm{An~MLP}$ \;
			
			Compute new link errors: $e_{uv}^{\left \langle t \right \rangle} = {R_v^{\left \langle t \right \rangle}}^{-1} \Tilde{R}_{uv} R_u^{\left \langle t \right \rangle}$ \;
			
			Compute absolute rotation loss: $\mathcal{L}_{a}^{\left \langle t \right \rangle}\left( v\right) = d_Q \left ( \hat{R}_v^{\left \langle t \right \rangle}, {R_v}^{\left \langle t \right \rangle}\right) \quad \forall v \in V$ \;
    		
    		Compute relative rotation loss: $\mathcal{L}_{r}^{\left \langle t \right \rangle} \left( v, u\right) = d_Q \left ( (\hat{R}_v^{\left \langle t \right \rangle})^{-1}\hat{R}_u^{\left \langle t \right \rangle}, (R_v^{\left \langle t \right \rangle})^{-1} R_u^{\left \langle t \right \rangle} \right)  \quad \forall (u,v) \in \epsilon$ \;
    		
    		Compute iteration loss: $\mathcal{L}^{\left \langle t \right \rangle} = \frac{1}{\left | V \right |} \sum_{v \in V} \left (\frac{1}{\left |\mathcal{N}_v  \right |}\sum_{u \in \mathcal{N}_v} \mathcal{L}_{r}^{\left \langle t \right \rangle}\left( v, u\right) + {\beta} \mathcal{L}_{a}^{\left \langle t \right \rangle}\left( v\right)  \right ) $ \;
			
			$t \gets t + 1$ \;
		}
        Compute overall loss: 
		Update all MLP weights using back propagation \;
		\caption{Learning in Attention based rotation averaging. Batch size equal 1 graph.}
	\end{algorithm*}
	
	\subsection{Message Aggregation}
	Two aspects of the rotation averaging problem led to the next contribution, the graph initialization technique and the SFM graph topology. 
	Nodes far from the origin have been initialized by a path involving many relative rotations or edges between nodes. These are more likely to have poor initialization due to the accumulated error in the relative rotations and the increased probability of an outlier or corrupted edge existing in the initialization path. These nodes far from the origin, are more difficult to average as they are generally sparsely connected with fewer neighbors and therefore less messages are being propergated for these nodes.\\
	
	With these sparsely connected nodes making up the majority of the outliers after rotation averaging is performed, focus was put on getting more reliable messages from the center of the graph to the outskirts where these problem nodes lie. Intuitively the method of selecting an origin suggest that central nodes have less error both before and after rotation averaging is performed . A simple measure of this centrality and reliability of a nodes value is the number of predictions available for that node on a given iteration, i.e., the number of neighbors a node has.
	
	\begin{figure}[h]
		\includegraphics[width=.5\textwidth]{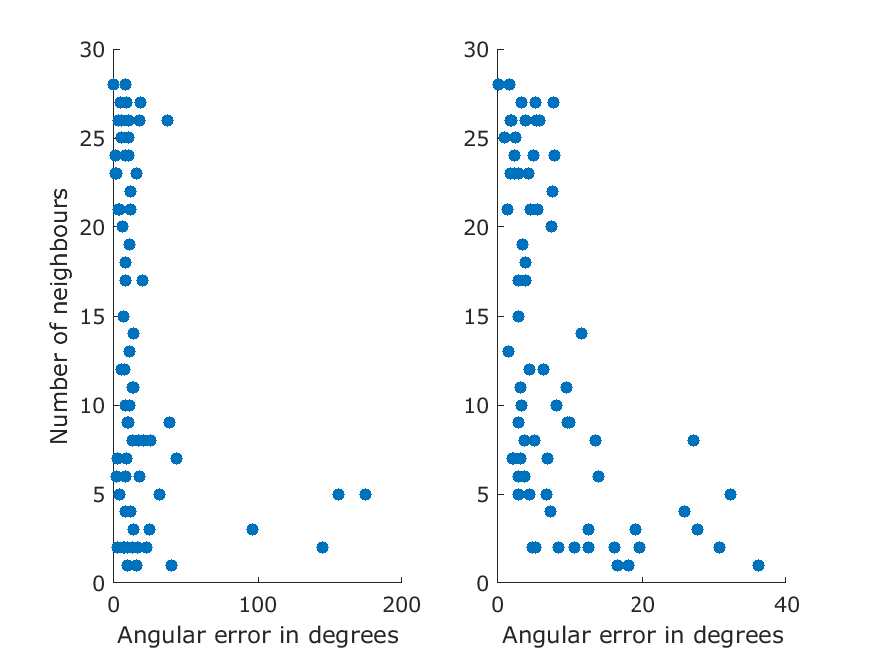}
		\caption{Scatter plot of a nodes angular error vs the number of neighbors the node has, in a small synthetic dataset before (left) and after (right) rotation averaging is performed}
		\label{fig:neighbors_scatter_plot}
	\end{figure}
	
	Plotting node angular error against the number of neighbors, visually showed the inverse correlation between these parameters \ref{fig:neighbors_scatter_plot}. Changing the message aggregation method from the mean aggregation used in \cite{Gilmer2017, Purkait2019}, to a weighted average with weights proportional to the number of neighbors a source node has, harnesses this idea of graph centrality and helps propagate messages from the center where nodes are densely connected, toward the outskirts. The results of using weighted averaging aggregation improved over mean aggregation; however, this puts trust entirely on source nodes, ignoring the fact that edges may be untrustworthy and could corrupt a message during the propagation phase.

	\subsection{Edge Attention}
	One of the problems with solely relying on a number of neighbors based weighted averaging is that the relative rotation between two nodes may be corrupted. If the source node has many neighbors and the weighting is high, a corrupted edge would result in an amplification of a poor message propagation. 	
	For this reason the number of neighbors based weighted averaging is coupled with an edge attention based model. \\
	Recent attention-based graph methods \cite{Hong2019, Velickovic2017, Ryu2018} rely on techniques that change the weighting on the edge messages during propagation depending on graph constraints, properties or assumptions. In chemical applications such as \cite{Hong2019,Ryu2018,Shang2018} attention is given differently for various edge attributes. Natural language processing is another common application of attention models and methods such as \cite{Guo2019} use word-level and sentence level attention to give weighting depending on the types of words encountered and where they are positioned in the sentence. \\
	
	Rotation averaging graphs do not have different types of edges or nodes and the graph topology is not predictable or known. The only source of information for attention based edge adaptiveness in rotation averaging is the edge features themselves. \cite{Gong2018} harnessed this view by adding attention coefficients and functions to the edge features that got passed between layers. \\
	
	The proposed attention model, weighs each edge message based on edge features and the source nodes neighbor information. This can be described by
	\begin{equation}\label{eq:w_uv}
		{w}_{uv}^{\left \langle t \right \rangle} = \frac{\exp \left (\hat{w}_{uv}  \right )}{\sum_{j\in \mathcal{N}_v} \exp \left (\hat{w}_{jv}  \right )}
	\end{equation}
	\begin{equation} \label{eq:w_uv}
		\hat{w}_{uv} = f_a\left(\frac{\left |\mathcal{N}_u  \right |}{\max_{\{j \in \mathcal{N}_v\}} |\mathcal{N}_j |}, e_{uv}^{\left \langle t-1 \right \rangle} \right)
	\end{equation}
	where $f_a$ is a learnable function that learns which messages to pay more attention to by using the number of neighbors intuition and edge features to determine the message reliability.
	\section{Results}
	\begin{table*}[t]
		\small
		\tabcolsep=0.1cm
		\begin{center}
			\begin{tabular}{|ccccc|ccccc|ccccc|}
				\hline
				\multicolumn{5}{|c}{Chatterjee} & \multicolumn{5}{|c}{NeuRoRA} & \multicolumn{5}{|c|}{Ours} \\
				\hline
				mean& med& RMS& \%\textgreater10$^{\circ}$& \%\textgreater30$^{\circ}$& mean& med& RMS& \%\textgreater10$^{\circ}$& \%\textgreater30$^{\circ}$&  mean& med& RMS& \%\textgreater10$^{\circ}$& \%\textgreater30$^{\circ}$\\
				7.129&	4.214&	10.7&	18.31&	4.225&	6.348& 2.828&	11.583&	15.493&	5.634&	5.608&	3.417&	7.78&	18.31&	0\\
				\hline
			\end{tabular}
		\end{center}
		\caption{Comparison of MRA performed on a synthetic dataset state-of-the-art methods.}
		\label{tab:synth}
	\end{table*}
	\subsection{Training}
	The dual sequential networks used by \cite{Purkait2019} make the training process slow and complicated. The first network requires training using an uninitialized dataset, then inferences from this network are initialized with the original data to generate the training data for the second network. This makes it very difficult to get any insights into the training performance of each network. No conclusive performance metrics can be generated until both networks are trained so the first network is trained blind, whilst the performance of the second network is largely dependent on the performance of the first.\\
	Our method uses a single network that does not require an intermediate initialization stage. Our network also does not require outliers to be calculated and provided as part of the ground truth for training. It also does not remove any nodes from the output so the input and output graph topology is identical. This simplifies the training process and reduces the barrier to entry for neural based rotation averaging.
	
	\subsection{Performance Metrics}
	\subsubsection{Accuracy}
	There is no standard method to compare predicted output graphs to the ground truth; as such, previous works have been inconsistent in their comparison methods. Before two graphs can be compared, they need to share a common frame of reference.
	The method used by\cite{Purkait2019} sets the most connected node in the ground truth as the origin and calculates the transform required to make the most connected node in the prediction equal to that of the ground truth. The error of the most connected node is therefore zero as it matches the ground-truth exactly; however, with this being the only node that is aligned, the rest of the graph alignment is heavily reliant on the accuracy of the prediction for this particular node. A more robust approach as used in this work, was discussed by \cite{Chatterjee2017} where the geodesic distance between all corresponding nodes in the two graphs is minimized. There is also debate over the loss function used to minimize the distances. Commonly, works minimize the sum of absolute errors but this has a tendency to produce favorable results to the median and mean performance metrics even when graphs have not been averaged well. We have chosen to use the sum of squares loss function as it is more robust to portions of the graph with poor predictions and will better highlight this in the  performance metrics.
	
	To get a complete idea of how well the rotation averaging has been performed, the geodesic distance between the nodes of the prediction and ground truth are compared. Statistics such as mean and median are commonly used to compare methods, but they don't capture nearly enough of how well a graph has been averaged. Additional metrics such as percentage of nodes that have errors above intervals such as 10, 15, 30, 60, 90 deg can expose how robust methods are to outliers in a dataset.
	
	\subsubsection{Speed}
	Speed is another important metric that is compared. It is difficult and often unfair to compare speeds as previous works are written in different languages, with implementations that harness hardware differently. \cite{Purkait2019} was the first to produce a rotation averaging method that could theoretically run in real time. Our method has improved on speed further and reduced the need to manually transfer data between different programs. Both ours and \cite{Purkait2019} utilize deep learning frameworks that have been highly optimized for parallel computing. This can often be orders of magnitude faster with a CPU only comparison between multi-threaded deep learning frameworks and single threaded traditional methods and therefore comparing speed is somewhat pointless. The extremely fast inference time of the deep learning frameworks often increases another order of magnitude or two when utilizing GPU acceleration, further increasing the speed difference between it and conventional optimization algorithms.  The disadvantage of deep learning methods are that they require a lot of time in the data acquisition and training process which is not required for conventional optimization methods to be deployed.
	
	\begin{table}[h]
	\centering
        \begin{tabular}{clll}
        \hline
        \multicolumn{1}{l}{\textbf{}} &  & \multicolumn{2}{c}{Inference Time (s)} \\ \hline
        \multicolumn{1}{l}{\# Cameras} & Method & CPU & GPU \\ \hline
        \multirow{3}{*}{100} & Chatterjee & 0.033 & NA \\
         & NeuRoRA & 0.088 & 0.014 \\
         & Ours & \textbf{0.013} & \textbf{0.012} \\ \hline
        \multirow{3}{*}{400} & Chatterjee & 0.96 & NA \\
         & NeuRoRA & 0.11 & 0.018 \\
         & Ours & \textbf{0.081} & \textbf{0.014} \\ \hline
        \multirow{3}{*}{1000} & Chatterjee & 11.1 & NA \\
         & NeuRoRA & 0.58 & 0.046 \\
         & Ours & \textbf{0.45} & \textbf{0.043} \\ \hline
        \end{tabular}
        \caption{Comparison of runtimes for synthetic datasets of various sizes}
        \label{tab:syn_runtime}
    \end{table}
	Table \ref{tab:syn_runtime} shows a comparison of the runtime speeds of various methods, for graphs with a varying number of cameras. The conventional iterative method offered by Chatterjee is comparable to the neural methods for a small number of cameras. As the number of cameras increases the difference between the conventional iterative optimization methods and the neural frameworks become orders of magnitude. It is also worth noting that the neural methods runtime was taken as the inference time only. There is time required to load models into memory; however, this only needs doing once prior to any number of predictions being infered. Furthermore NeuRoRA requires the SPT initialization to be run in a separate program between the inference of the first and second networks, where the other methods run directly on the SPT initialized graphs.
	\subsection{Synthetic Datasets}
	Synthetic datasets were created that are representative of the typologies, noise and outlier distributions seen in real datasets as shown by \cite{Purkait2019}.
	The comparative results for our method on a synthetic dataset can be seen in table \ref{tab:synth}.

	\section{Conclusions}
	We have presented a new deep learning approach to the problem of multiple rotation averaging. We have overcome the barriers to entry of the recent neural based approach by using a single network that can robustly perform rotation averaging without the need for a separating cleaning network. This new approach recognizes that messages being passed within the neural message passing framework contain outliers and performing mean aggregation is not robust enough to give accurate predictions for next layer node-level hidden representations.
	The solution to overcome this is to provide a message attention mechanism that will score message reliability and then pool the messages, giving more weight to statistically more reliable messages and less to messages corrupted by outliers.
	The result of this robust message attention mechanism is that all observations are considered, with outliers affect on the results supressed, rather than the outliers being identified and removed based on human assumptions.
	This proposed single network  has comparable accuracy results to previous state-of-the art methods whilst providing faster inference times and being significatly simpler and quicker to train than the previous learning based MRA method.

	{\small
		\bibliographystyle{ieee_fullname}
		\bibliography{v1}
	}
	
\end{document}